\documentclass{article}
\usepackage{ijcai22}

\usepackage{times}
\usepackage{soul}
\usepackage{url}
\usepackage[hidelinks]{hyperref}
\usepackage[utf8]{inputenc}
\usepackage[small]{caption}
\usepackage{graphicx}
\usepackage{amsmath}
\usepackage{enumerate}
\usepackage{bm}
\usepackage{amsthm}
\usepackage{booktabs}
\usepackage{algorithm}
\usepackage{algorithmic}
\usepackage{nicefrac}
\usepackage{mathtools}
\usepackage[dvipsnames]{xcolor}
\usepackage{amsfonts}

\newcommand{\cA}{\mathcal{A}}

 \newcommand{\cL}{\mathcal{L}}
\newcommand{\cM}{\mathcal{M}}

\newcommand{\cK}{\mathcal{K}} \newcommand{\cY}{\mathcal{Y}}
 \newcommand{\cX}{\mathcal{X}}

\newcommand{\EE}{\mathbb{E}} \newcommand{\RR}{\mathbb{R}}

\newtheorem{problem*}{Problem}

\DeclareMathOperator*{\argmin}{argmin}

\newcommand{\Var}{\mathrm{Var}}

\usepackage{todonotes}
\usepackage{setspace}

\title{Differential Privacy and Fairness in Decisions and Learning Tasks: A Survey}

\author{
Ferdinando Fioretto$^1$\footnote{Authors order is alphabetical. All authors contributed equally.}\and
Cuong Tran$^1$\and
Pascal Van Hentenryck$^2$\And
Keyu Zhu$^{2}$
\affiliations
$^1$Syracuse University\\
$^2$Georgia Institute of Technology
\emails
\{ffiorett, cutran\}@syr.edu,
pvh@isye.gatech.edu,
kzhu67@gatech.edu
}

\begin{document}

\maketitle

\begin{abstract}
This paper surveys recent work in the intersection of differential privacy (DP) and fairness. 
It reviews the conditions under which privacy and fairness may have aligned or contrasting goals, analyzes how and why DP may exacerbate bias and unfairness in decision problems and learning tasks, and describes  available mitigation measures for the fairness issues arising in DP systems. The survey provides a unified understanding of the main challenges and potential risks arising when deploying privacy-preserving machine-learning or decisions-making tasks under a fairness lens.
\end{abstract}

\section{Introduction} The availability of large datasets and computational resources has driven significant progress in Artificial Intelligence (AI) and, especially, Machine Learning (ML). These advances have rendered AI systems instrumental for many decision making and policy operations involving individuals: they include
assistance in legal decisions, lending, and hiring, as well determinations of resources and benefits,
 all of which have profound social and economic impacts. While data-driven systems have been successful in an increasing number of tasks, the use of rich datasets, combined with the adoption of black-box algorithms, has sparked concerns about how these systems operate. How much information these systems leak about the individuals whose data is used as input and how they handle biases and fairness issues are two of these critical concerns.

\begin{figure}[!t]
\centering
\includegraphics[width=\linewidth]{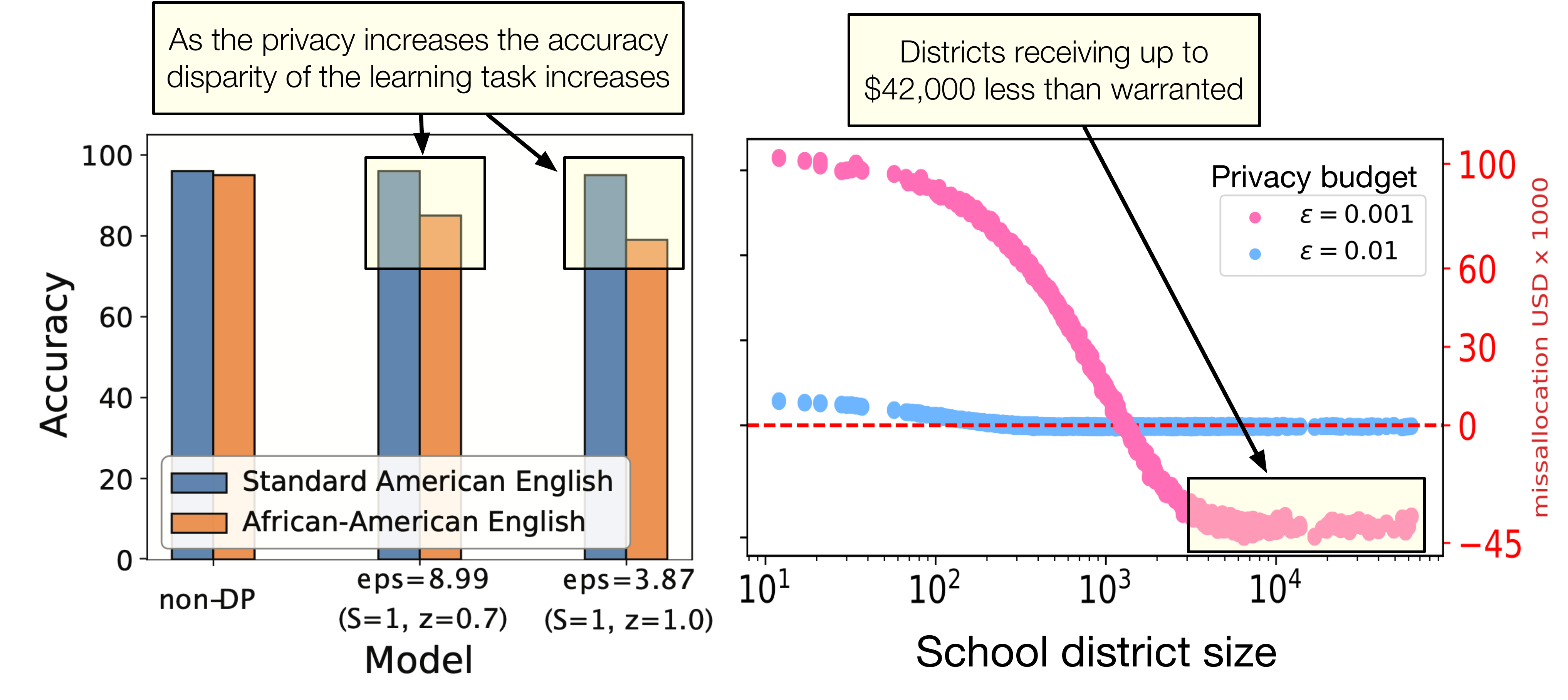}
\caption{Left: Disparities arising in DP sentiment analysis tasks (image from \protect\cite{NEURIPS2019_eugene}). 
Right: Disparity arising in fund allocations to school districts (image from \protect\cite{tran2021decision}).}
\label{fig:motivation}
\end{figure}

\emph{Differential Privacy} (DP) \cite{dwork:06} has become the paradigm of choice for protecting data privacy and its deployments are also growing at a fast rate. These include several data products related with the 2020 release of the US.~Census Bureau \cite{abowd2018us}, and
by Google \cite{aktay2020google}, Facebook \cite{facebook}, and Apple \cite{apple}. DP is
appealing as it bounds the risks of disclosing sensitive information for individuals participating in a computation. 
However, the process adopted by a DP algorithm to ensure the privacy guarantees involves calibrated perturbations, which inevitably introduce errors to the outputs of the task at hand. More importantly, 
it has been shown that these errors may affect different groups of individuals differently. 
An example of these effects are reported in Figure \ref{fig:motivation} (left), which illustrates that a DP learning model affects the accuracy of 
the minority group (African-American) more than it does the majority group in a sentiment analysis of Tweets \cite{NEURIPS2019_eugene}. Similar observations were reported in decision tasks (Figure \ref{fig:motivation}, right) in which privacy-preserving census data is used to allocate funds to school 
districts \cite{pujol:20,tran2021decision}.
The illustration shows that, under a privacy-preserving allocation scheme, some school districts may systematically receive considerably less money than 
what would be warranted otherwise.

These effects can have significant societal and economic impacts on the involved individuals: classification errors may penalize some groups over others in important determinations, including criminal assessment, landing, and hiring, or can result in disparities regarding the allocation of critical funds, benefits, and therapeutics. 
These fairness issues in DP settings are receiving increasing attention, but a
complete understanding of why they arise is
still limited. For example, it is often believed that post-processing the output of a differential private data-release mechanism may introduce bias and reduce errors  but the underlying phenomena have only recently started to be studied in detail. Furthermore, in privacy-preserving learning tasks, it is (often incorrectly) believed that disparate impacts are caused by the presence of unbalanced data. 
{\em It is the goal of this survey to demystify some common
beliefs about the interaction between differential privacy and fairness, and provide a critical review of the state of knowledge in this important area.} 
The survey focuses on two key privacy-preserving processes: \emph{downstream decisions tasks}, in which a privacy-preserving version of a sensitive dataset is used to allocate resources or grant benefits, and \emph{learning tasks}, in which a learning model is rendered differentially private.

\section{Preliminaries}
\label{sec:preliminaries}
This section reviews the notion of differential privacy and compares some 
key fairness concepts adopted in this survey. 

\smallskip\noindent\textbf{Differential Privacy} (DP) \cite{dwork:06} is 
a rigorous privacy notion that characterizes the amount of information 
of an individual's data being disclosed in a computation.
A randomized mechanism $\cM:\cX \to \cY$ with domain $\cX$ and range 
$\cY$ satisfies $(\epsilon, \delta)$-\emph{differential privacy} if, for 
any output $y \in \cY$ and datasets $\bm{x}, \bm{x}' \in \cX$ 
differing by at most one entry, 
  \begin{equation}
  \label{eq:dp}
    \Pr[\cM(\bm{x}) = y] \leq \exp(\epsilon) \Pr[\cM(\bm{x}') = y] + \delta.
  \end{equation}

\noindent
Intuitively, DP states that outputs to the privacy-preserving mechanism are returned
with a similar probability regardless of whether the dataset includes a specific
individual.  Parameter $\epsilon > 0$
describes the \emph{privacy loss} of the mechanism, with values close
to $0$ denoting strong privacy. When $\delta = 0$, mechanism $\cM$ is
said to be $\epsilon$-differentially private.
Differential privacy satisfies several important properties. Notably, \emph{composability} ensures that a combination of DP mechanisms preserves  differential privacy and 
\emph{post-processing immunity} ensures that privacy guarantees are
preserved by arbitrary data-independent post-processing steps \cite{Dwork:13}.

\paragraph{Fairness}
This survey focuses on two main fairness notions: \emph{individual} and \emph{group}
fairness. 
Individual fairness \cite{dwork2012fairness} claims that \emph{similar 
individuals should be treated similarly}. For a mechanism $\cM$ mapping inputs
$\bm{x} \in \cX$ to outputs $y \in \cY$, individual fairness is satisfied
when for any $\bm{x}, \bm{x}' \in \cX$:
\begin{equation}
 d_{\cY}(\cM(\bm{x}), \cM(\bm{x}')) \leq d_{\cX}(\bm{x}, \bm{x}'), 
 \label{eq:ind_fair_def}
\end{equation}
where $d_\cX \!:\!\cX \times \cX \to \RR_{+}$ and $d_\cY\!:\! \cY \times \cY \to \RR_{+}$, 
are distance metrics over pairs of inputs and outputs, respectively.
When condition \eqref{eq:ind_fair_def} holds, mechanism $\cM$ is said to satisfy a 
$(d_{\cX}, d_{\cY})$-Lipschitz condition.
An obvious drawback of individual fairness is its requirement of 
problem-specific distance metrics, which may not be easy to 
design.

Group fairness, in contrast, requires some statistical property of 
any protected group of individuals (e.g., a defined by gender or race)  to be similar to that of the whole population. 
Examples of commonly adopted group fairness notions are 
\emph{demographic parity}, which is satisfied when the outputs of a predictor $\cM$ are statistically independent of the protected group attribute \cite{dwork2012fairness}, 
\emph{equal opportunity}, which is satisfied when $\cM$'s predictions are conditionally independent of the protected group attribute for a given label \cite{hardt2016equality}, and 
\emph{accuracy parity}, which is satisfied when $\cM$'s miss-classification 
rate is conditionally independent of the protected group attribute \cite{zhao2019inherent}. A connection between group fairness and individual fairness was presented 
by Dwork et al.~\shortcite{dwork2012fairness}, who showed that if a model is individually fair 
and if the Earthmover distance across the protected groups data is sufficiently 
small, then such model also satisfies demographic parity. 

\begin{figure}[!tb]
\centering
\includegraphics[width=1.05\linewidth]{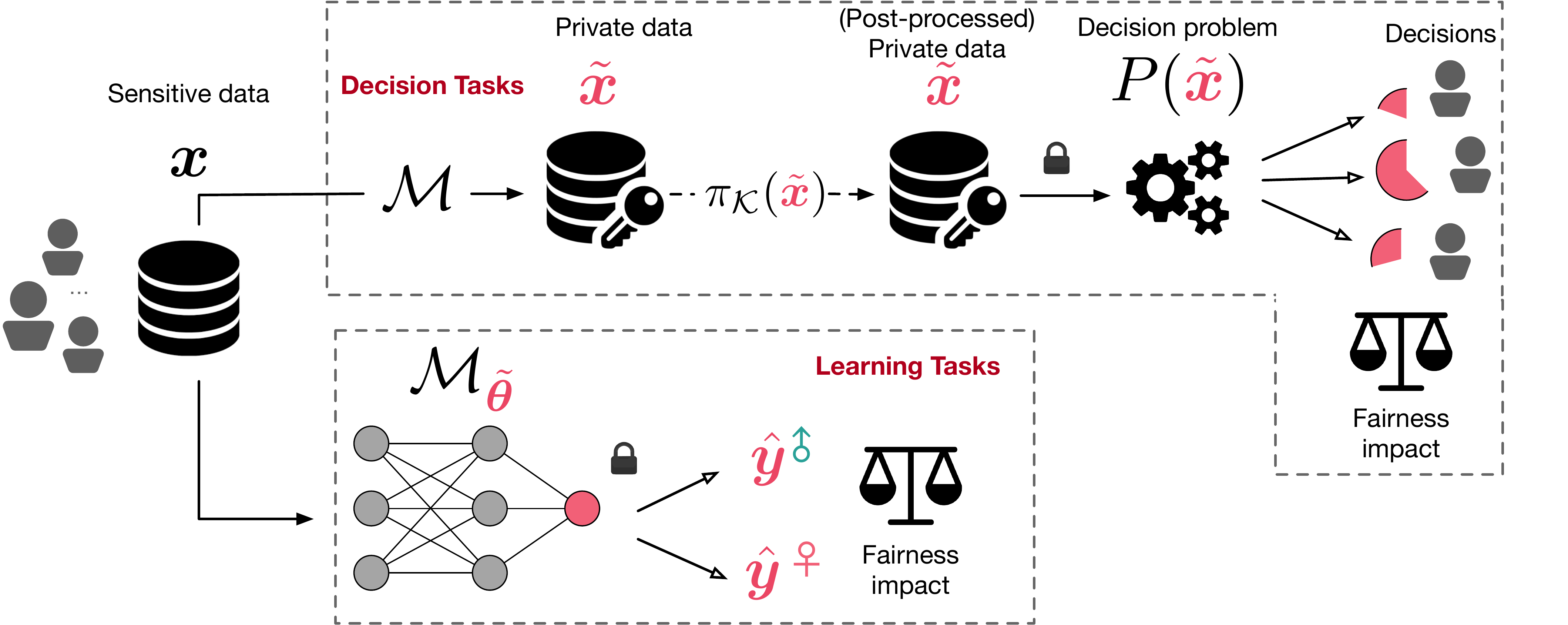}
\caption{Setting analyzed in this survey.}
\label{fig:setting}
\end{figure}

\section{Settings}
\label{sec:settings}

The focus of the survey is to shed light on the disproportionate effects 
induced by a DP mechanism $\cM$ on the outputs of some 
task of interest. The  considered mechanisms process inputs 
$\bm{x} \in \cX$ of $n$ entries, containing sensitive information, 
such as the individuals' ethnicity, salary, gender, and geographic locations. 
Within this setting, the survey focuses on \emph{decision tasks} and 
\emph{learning tasks}, whose schematic illustrations are shown in Figure \ref{fig:setting}.

\paragraph{Decision Tasks} 
This setting considers data-release mechanisms $\cM$ producing a 
privacy-preserving counterpart $\tilde{\bm{x}}$ of $\bm{x}$. Then
the DP dataset $\tilde{\bm{x}}$ is used as the input to a decision problem 
$P \!:\! \cX \!\to\! \cY \subseteq \mathbb{R}$. 
For instance, $P$ may describe an allotment of funds to school districts. 
{This setting is widely adopted in several data-release tasks, 
including census applications and allocation of renewable energy 
resources in energy markets.}
Mechanism $\cM$ may also apply a post-processing
step $\pi_\cK$ to restrict the randomized output $\tilde{\bm{x}}$ to
be within a feasible region $\cK$, e.g., to guarantee non-negativity
of the released data.
The focus of this task is to study the effects of a DP data-release 
mechanism $\cM$ to the outcomes of problem $P$ in relation to the fairness 
of the decisions. 
Because random noise is added to the original dataset $\bm{x}$, the
output $P(\tilde{\bm{x}})$ incurs some error. A quantification of the disparate 
impact of this error among the problem entries is often measured through the 
bias of problem $P$ for some entry $i \!\in\! [n]$,
\begin{equation}
 \label{eq:bias}
  B_P^i(\cM, \bm{x}) = 
  \EE_{\tilde{\bm{x}} \sim \cM(\bm{x})} \left[ P_i(\tilde{\bm{x}}) \right] - P_i(\bm{x}),
 \end{equation}
 where $P_i$ is used to denote the program computing the output associated with entity $i \in [n]$. This bias characterizes the distance between the expected
privacy-preserving decision and the decision obtained on the real
data (ground truth). In this context, the fairness analysis attempts to bound the maximal difference in bias among any pairs of entries: 
$\max_{i,j \in [n]} |B_P^i(\cM, \bm{x}) - B_P^j(\cM, \bm{x})|$.

\paragraph{Learning Tasks}
In this second setting, $\cM_{\tilde{\theta}}$ is a classifier parametrizded by vector $\tilde{\theta}$ that protects the disclosure of the individuals in $\bm{x}$ and the focus is to analyze 
the fairness impact of privacy on different groups of individuals. 
The elements of $\bm{x}$ are data points $(x, a, y)$ where 
$x \!\in\! \cX$\footnote{Used here to denote the feature set, slightly abusing notation.} is a feature vector, $a \!\in\! \mathcal{A}$ is a protected group 
attribute, and $y \!\in\! \mathcal{Y}$ is a label.
The model quality is measured by a \emph{loss function} 
$\ell: \mathcal{Y} \times \mathcal{Y} \to \mathbb{R}_+$, and the problem is 
to minimize the empirical risk function:
\(
\label{eq:erm}
    \min_\theta \cL(\theta; \bm{x}) = \EE_{(x,a,y)}
    [\ell(\cM_\theta(x), y)].
\)
Methods reviewed in this survey analyze the disparate impact of privacy on different groups of individuals either by measuring the deviation from a model to satisfy a notion of group fairness exactly or 
using the notion of \emph{excessive risk} \cite{zhang2017efficient,pmlr-v97-wang19c}. The latter  defines the difference between the private and non private risk functions: 
\begin{flalign}
\label{def:excessiver_risk}
  R(\theta, \bm{x}_a) = \EE_{\tilde{\theta}}  
  \left[ \cL( \tilde{\theta}; \bm{x}_a) \right]
       - \cL( \theta^*; \bm{x}_a),
\end{flalign}
where the expectation is defined over the randomness of the private mechanism,
$\bm{x}_a$ denotes the subset of $\bm{x}$ containing exclusively samples whose group attribute is $a$, $\tilde{\theta}$ denotes the private model parameters, and $\theta^* = \argmin_\theta \cL(\theta; \bm{x})$.
In this context, (pure) fairness is achieved when there is no difference in excessive risk across all protected groups.

\section{Privacy and Fairness: Friends or Foes?}
\label{sec:friends_of_foes}
While DP aims at rendering the participation of individuals indistinguishable to an observer who accesses the outputs of a computation, fairness attempts at equalizing properties of these outputs across different individuals. Thus, simultaneously achieving these two goals has received two contrasting views. The first sees privacy and fairness as 
\emph{aligned} objectives while the second sees them as \emph{contrasting} ones. 

Contributions in the "aligned space" focus on studying conditions 
for which privacy and fairness can be achieved simultaneously. Notably, 
Dwork et al.~\shortcite{dwork2012fairness} shows that individual fairness is a generalization 
of differential privacy. 
To see why privacy and fairness may be achieved simultaneously, notice 
that a mechanism $\cM:\cX \to \cY$ 
satisfies $\epsilon$-differential privacy when it is $(d_{\cX}, d_{\cY})$-Lipschitz with 
\begin{align*}
d_{\cX}(\bm{x}, \bm{x}') &= \epsilon |\bm{x} \Delta  \bm{x}'| \\
d_{\cY}(\cM(\bm{x}), \cM(\bm{x}')) &= \sup_{y \in \cY} \log 
\left(
\frac{\Pr(\cM(\bm{x}) = y)}{\Pr(\cM(\bm{x}') = y)}
\right),
\end{align*}
\noindent
where $ \bm{x} \Delta \bm{x}'$ represents the set difference between 
two inputs $\bm{x}$ and $\bm{x}'$ of $\cX$. 
Thus, DP mechanisms also ensure 
individual fairness, as long as $d_\cX$ and $d_\cY$ are defined as above.
Similarly, Mahdi et al.~\shortcite{mahdi2020improving} shows that, 
in candidate selection problems, the use of a DP exponential mechanism \cite{mcsherry2007mechanism} produces fair selections when the data satisfies some restrictions concerning key properties (average and variance of the qualification scores) of each group.

The second line of works views privacy and fairness as contrasting goals. 
Notably, it has been observed that the outputs of DP classifiers may create or exacerbate disparate impacts among groups of individuals 
\cite{NEURIPS2019_eugene}. A similar phenomenon was also reported in important decision tasks that use DP census statistics as inputs \cite{pujol:20}. These works typically adopt the notion of group fairness and impose no 
restrictions on the properties of the privacy-preserving mechanisms 
studied. The rest of the survey focuses on analyzing why these important observations
arise and how can they be mitigated.

\section{Why Privacy impacts Fairness?}
\label{sec:analysis}

This section reviews the current understanding about why disparate impacts arise in two common privacy-preserving processes: decision tasks and 
learning tasks.

\subsection{Decision tasks}
\label{subsec:decision_tasks}

Consider first a data-release mechanism $\cM$, which typically consists 
of two steps: First, noise drawn from a calibrated distribution is injected into the original data $\bm{x}$ to obtain a DP counterpart $\tilde{\bm{x}}$.
This process, however, may fundamentally affect some important properties 
of the original data. For example, if $\bm{x}$ is a vector of counts
enumerating individuals living in different regions, its privacy-preserving
version $\tilde{\bm{x}}$ may not satisfy non-negativity conditions. 
Thus, a post-processing step $\pi_\cK$ is applied to $\tilde{\bm{x}}$ 
to redistribute the noisy values in a way that the resulting outputs 
$\pi_{\cK}(\tilde{\bm{x}})$ satisfy the desired data-independent constraints $\cK$.
Second, the released data $\tilde{\bm{x}}$ is used as input to a decision 
problem $P$. This pipeline is shown in Figure \ref{fig:setting} (top).
The goal of this section is to characterize the disparity in errors 
induced by mechanism $\cM$ on the final decisions $P(\tilde{\bm{x}})$. 
 
The negative impacts of privacy towards fairness in decision tasks were
first observed by Pujol et al.~\shortcite{pujol:20}. 
The authors noticed that the use of privacy-preserving census data to allocate 
funds to school district produces unbalanced allocation errors, with 
some school districts systematically receiving more (or less) than 
what warranted, as illustrated in Figure \ref{fig:motivation} (right). 
A similar behavior was also observed in other census-motivated decision 
tasks, including determining whether a jurisdiction qualifies for 
providing minority language assistance during an election, and 
apportionment of legislative representatives.  

These empirical observations were later attributed to two main 
factors: (1) the ``shape'' of the decision problem $P$ \cite{tran2021decision} and 
(2) the presence of non-negativity constraints in post-processing steps \cite{zhu:20_postdp}.
The survey reviews next these two factors in details.

\paragraph{Shape of the decision problem.} 
Note that private data is often calibrated with unbiased
noise, such as Laplacian noise in the Laplace mechanism,
for privacy protection. In such contexts Tran et al.~\shortcite{tran2021decision} 
showed that a decision problem that applies a linear transform 
of its input yields an unbiased outcome with respect to the true outcome. 
However, non-linearities in the decision problem are likely to 
generate non-zero biases with discrepancies among entities, 
which results in fairness issues. 
In more details, when $P_i$ is at least twice differentiable, 
the problem bias can be approximated as 
{\small
\begin{align}
    \notag
  B^i_P(\cM, \bm{x}) &= \EE[P_i(\tilde{\bm{x}} = \bm{x} + \eta)] - P_i(\bm{x})\\
    \label{eq:p1}
  &\approx \frac{1}{2}\bm{H}P_i(\bm{x}) \times \Var[\eta]
\end{align}
}
\noindent
where $\bm{H}P_i(\cdot)$ denotes the Hessian of problem $P_i$. 
The approximation above uses a Taylor expansion of
the private problem $P_i(\bm{x}\! + \!\eta)$ and the linearity of
expectations, with $\eta$ a random variable following some symmetric distribution. 
The bias $B^i_P$ can thus be approximated by an expression
involving the local curvature of the problem $P_i$ and the variance of
the noisy input (which depends on the privacy loss $\epsilon$). In turn, fairness violations (the maximal difference of the bias between any two entries) can be bounded whenever the problem local curvature is constant across entities, since the variance is also constant and bounded. 

Observe that the fairness violations  are controlled by both the privacy loss value $\epsilon$ (appearing in the variance term) and the shape of the decision problem (appearing in the Hessian term). Tight privacy requirements (small $\epsilon$ values) or non-linearities in the decision problem may lead to large disparate impacts.
An important conclusion of this result is that using DP to generate private inputs of decision problems commonly adopted to make policy determinations {\em will necessarily introduce fairness issues, despite the noise being unbiased}.

Notice that the analysis above holds for problems with continuous support. 
In the case of \emph{Boolean decision functions}, e.g., $P_{i} : \cX \to \{0,1 \}$, it was shown that disparate impacts may be exacerbated when multiple decision functions are composed \cite{tran2021decision}. This is of particular interest in policy decisions like those used to determine whether a jurisdiction qualifies for a particular benefit, such as the minority language voting right problem \cite{pujol:20} where thresholding Boolean functions are ``concatenated'' using logical connectors. 
In a nutshell, the result shows that composing two decision problems with fairness violations bounded by values $\alpha_1$ and $\alpha_2$, respectively, produces a fairness violation bound $\alpha > \max(\alpha_1, \alpha_2)$.

\paragraph{Impact of post-processing}
Post-processing immunity is a fundamental property of differential
privacy which is routinely applied in many applications, including census data \cite{abowd2018us}, energy systems \cite{fioretto:TSG20}, and mobility \cite{xi:15,Fioretto:AAMAS-18}.
Notably, when the feasible region is convex, a largely adopted class of post-processing functions, called \emph{projections}, is guaranteed to improve accuracy \cite{HayRMS10,fioretto2021differential}. 

\begin{figure}[!t]
\centering
\includegraphics[width=0.85\linewidth]{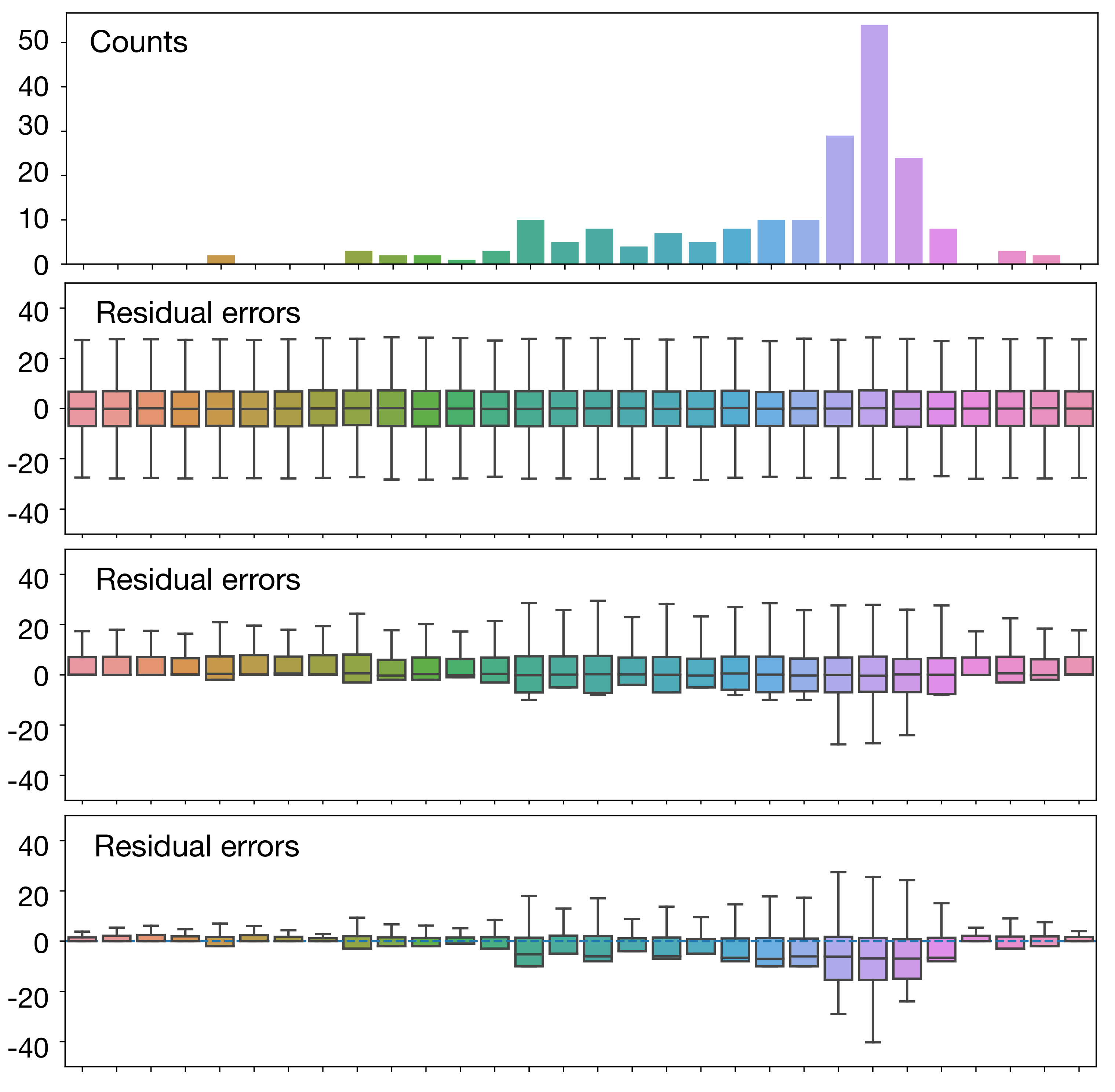}
\caption{{\small Bias and variance in DP post-processing.}}
\label{fig:postproc}
\end{figure}

While post-processing is often used to reduce errors, this step can
also introduce bias and fairness issues, as illustrated by Zhu et al.~\shortcite{zhu:20_postdp} and McGlinchey et al.~\shortcite{mcglinchey2020observations}. The issue is depicted in Figure \ref{fig:postproc}. It displays a histogram of population counts (top row) and the distribution of the residual errors $\tilde{\bm{x}} - \bm{x}$, where $\tilde{\bm{x}}$ is obtained by the application of Laplace noise on the true counts (second row). The third row displays the residual errors when applying
post-processing step 
$$\pi_{\geq 0 } \coloneqq \max(\bm{0}, \tilde{\bm{x}})$$ 
to enforce non-negativity. Finally, the fourth row is obtained by a constrained projection method 
{\small
$$ 
\pi_{\cK_S} \coloneqq \argmin_{\bm{v} \in \cK_S} 
\| \bm{v} - \tilde{\bm{x}}\|_2 \;,\; 
\cK_S = \{ \bm{v} \in \RR^n \;\mid\;\sum_i v_i = \tilde{S}, v_i \geq 0\},
$$} 
\hspace{-2pt}that enforces a linear constraint imposing that the sum of the
projected counts $\bm{v}$ be equal to a noisy sum \mbox{$\tilde{S} \!=\! 
(\sum_i x_i) \!+\! \eta$} with $\eta$ being some appropriately selected noise.  
All of the above private methods achieve the same privacy loss.  Note that the application of Laplace noise does not introduce bias: all the outputs have the same residual errors. However, even the simple 
non-negative post-processing step
produces bias and different error variances across the problem entries  \cite{mcglinchey2020observations}. 
The more complex mechanism $\pi_{\cK_S}$ further exacerbates the biases and variance differences as showed by Zhu et al.~\shortcite{zhu2022post}.
Additionally, Zhu et al.~\shortcite{zhu:20_postdp} showed that for the more general constraint spaces $\cK \!=\! \{\bm{x} \;\mid\; A \bm{x} \leq b, \bm{x} \!\geq\! 0\}$, the solution $ \pi_{\cK}(\tilde{\bm{x}})$ is an unbiased estimator of $\bm{x}$ when the non-negative constraint $\bm{x} \!\geq\! \bm{0}$ is ignored. However, when incorporating this non-negativity constraint, the optimal solution $ \pi_{\cK} (\tilde{\bm{x}})$ 
deviates from $\bm{x}$ statistically and, thus, resulting in non-zero bias.

 \subsection{Learning Tasks}
 \label{sec:issue_learning}
 
Consider now the disparate impacts arising in private learning tasks. 
These effects were first studied in the context of equal opportunity by  Cummings et al.~\shortcite{cummings2019compatibility}. They show that it is impossible to achieve pure fairness, i.e., 
$
\Pr[\cM(\bm{x}) \!=\! \hat{y} \!\mid\! a, y] \!=\! 
    \Pr[\cM(\bm{x}) \!=\! \hat{y} \!\mid\! y]
$, for all $a \!\in\! \cA$, 
when using an $\epsilon$-DP classifier. This result relies on the observation that a classifier $\cM$ cannot be perfectly fair for both a dataset $\bm{x}$ and a neighbor dataset $\bm{x}'$ differing from $\bm{x}$ by adding or removing one sample. 
However, a relaxed fairness goal, i.e., 
$ 
0 < | \Pr[\cM(\bm{x}) \!=\! \hat{y} \mid a, y] - 
    \Pr[\cM(\bm{x}) \!=\! \hat{y} \mid y] |
$
can be achieved by using the exponential mechanism, which satisfies $\epsilon$-DP \cite{cummings2019compatibility}. 

The effects of privately training deep learning models to the accuracy parity was first observed by Bagdasaryan et al.~\shortcite{NEURIPS2019_eugene}. The authors studied the disparities induced by DP Stochastic Gradient Descent (DP-SGD) \cite{AbadiCGMMT016}, the de-facto standard algorithm used to train deep learning models privately. They observed that the accuracy of the minority group was disproportionately impacted by the private training. These observations were validated on several vision and natural language processing tasks and in both a centralized and federated setting. 
The authors postulated that the size of a protected group would play a crucial role to the exacerbation of the disparate impacts in private training.
This behavior was also observed in a further empirical study which compared DP-SGD and PATE \cite{DBLP:conf/iclr/PapernotSMRTE18}, a popular semi-supervised DP learning framework. Therein, the authors reported PATE to cause milder disparate impacts when compared to DP-SGD under similar privacy constraints \cite{uniyal2021dp}. 
The hypothesis considering the size of a protected group as a predominant factor inducing the disparate impacts in private training was challenged by Farrand et al.~\shortcite{farrand2020neither}. The authors showed that the privacy preserving models can introduce substantial fairness issues even when slightly imbalanced datasets are considered. 

More recently, Tran et al.~\shortcite{tran2021differentially,tran2021fairness} show that group sizes may indeed not be a predominant factor to explain the disparate impacts observed in private training. The authors report two main factors contributing to these effects: {\bf (1)} the properties of the training data and {\bf (2)} the model’s characteristics. The survey reviews these factors next. 

\paragraph{Properties of the Training Data}
As observed by Tran et al.~\shortcite{tran2021differentially}, \emph{input norms} and \emph{distance to decision boundary} are two key characteristics of the data connected with exacerbating the disparate impacts of private learning tasks. 
First, by using an analysis analogous to that used in Equation \eqref{eq:p1}, it was shown that, when training convex models using output perturbation--which adds Gaussian noise to the output of the optimal models parameters--groups of samples associated with large Hessian losses $\bm{H}\ell(\theta; \bm{x}_a)$ can be penalized more than those associated with small Hessian losses. 
In turn, the authors show that groups with large input norms (often observed at the tail of the data distribution) may lead to large Hessian loss values. The observation that samples at tail of the data distribution are often penalized more than others was also reported by Bagdasaryan et al.~\shortcite{NEURIPS2019_eugene} and Suriyakumar et al.~\shortcite{DBLP:conf/fat/SuriyakumarPGG21}. 
Additionally, Tran et al.~\shortcite{tran2021differentially} show that the distance of a sample to the model decision boundary is also connected to the Hessian values. Samples which are near (far) to the decision boundary are less (more) tolerant to perturbations induced by the DP algorithm. This is intuitive, since perturbing the model parameters is more likely to impact samples which are close to the decision boundary. 
Similar observations were also reported in the context of DP-SGD \cite{tran2021differentially} and PATE \cite{tran2021fairness}. 

\paragraph{Model Characteristics}
In addition to the data properties, the characteristics of DP learning mechanisms have also been found connected with the disparate impacts of the private models. 
For example, at each training iteration, DP-SGD operates by computing the gradients for each data sample in a random mini-batch, clipping their L2-norm, adding noise to ensure privacy, and computing the average. 
The two key characteristics of DP-SGD are clipping the gradients whose L2 norm exceeds a given bound $C$ and perturbing the averaged clipped gradients with Gaussian noise. As shown in \cite{tran2021differentially}, both factors exacerbate unfairness in the private predictions of DP-SGD.
When different groups of individuals produce updates with large differences in magnitude of gradients norms, and when such values exceed the clipping bound $C$, then gradient clipping induces dissimilar information losses in these groups, thus penalizing those groups with larger gradients. This aspect was also observed in \cite{xu2021removing}. 
Additionally, the process of adding noise in DP-SGD is shown to produce an effect similar to that produced by the output perturbation reviewed above: The groups with larger Hessian losses defined on their samples tend to have the larger disparate impacts \cite{tran2021differentially}. 

Another important private ML framework is the \emph{Private Aggregation of Teacher Ensembles (PATE)} \cite{DBLP:conf/iclr/PapernotSMRTE18}. It combines multiple learning models used as teachers for a student model that learns to predict an output chosen by noisy voting among the teachers. The resulting model satisfies differential privacy and has been shown effective in learning high quality private models in semisupervised settings 
\cite{malek2021antipodes}. 
A key aspect of PATE is the scheme adopted by its teachers ensemble to privately predict labels which are used to train the student model. Tran et al.~\shortcite{tran2021fairness} showed that both the size of this ensemble and the confidence of the voting teachers are key factors in the analysis of the disparate impacts observed in this framework. 
Their analysis indicates that larger ensembles correspond to more robust predictions since the voting scheme becomes more consistent, given the noise added to guarantee privacy.

\section{Mitigating Fairness in Private Tasks}
\label{sec:solutions}

Having discussed the reason why disparate impacts arise in differentially private decision making and learning tasks, the survey reviews next the strategies proposed in the literature to mitigate the fairness issues arising in these two settings. 

\subsection{Decision tasks}
Several solutions have recently been developed to reduce the disparate impacts arising in DP decision tasks, with particular focus on a class of census-motivated problems used to distribute funds or grants benefits to the problem entities.
In particular, in the context of funds allocation to school district, Pujol et al.~\shortcite{pujol:20} proposed a mechanism which distributes additional budget to targeted entities, so that, all of the entities receive at least what warranted in a non-private allocation, with high probability. A limitation of this strategy is that the resulting allocation does not necessarily ensures feasibility (e.g., the sum of the individual allocations may not match a preassigned budget).
In the attempt to mitigate fairness issues for an analogous classes of problems, Tran et al.~\shortcite{tran2021decision} proposed to design a proxy problem that closely approximates the original decision task but admits bounded fairness. In particular, the authors observe that a linear approximation for an important class of allocation problems can be obtained when additional aggregated data can be released. 

Solutions to mitigate the disparate effects of post-processing have also been proposed. Of particular relevance is the work of Zhu et al.~\shortcite{zhu2022post}. The authors analyzed the fairness impact of projection mechanisms on a simplex and proposed a near-optimal projection operator which meets the feasibility requirements of allotment problems while providing substantial improvements in terms of disparate impact under different fairness metrics.

\subsection{Learning Tasks}
Mitigation strategies have also been designed in the context of learning tasks. 
Xu et al.~\shortcite{xu2019achieving} and Ding et al.~\shortcite{ding2020differentially} proposed versions of a fair and $\epsilon$- and $(\epsilon, \delta)$-DP logistic regression classifiers \cite{DBLP:journals/pvldb/ZhangZXYW12}. 
Both works target demographic parity and use a functional mechanism--which approximates the objective function of the classifier by a polynomial and injects calibrated noise to its coefficients. 

Most of the work attempting to mitigate the disparate impacts of DP in learning tasks has focused on the popular DP-SGD framework. 
DP-SGD does not restrict focus on convex loss functions rendering it an appealing framework for DP learning tasks. As observed in Section \ref{sec:issue_learning}, individual gradient clipping is a key factor in exacerbating the disparate impacts. Thus, Xu et al.~\shortcite{xu2021removing} proposes to associate a different clipping bound to each protected group, so as to limit the effect of disproportionate gradient pruning for those groups of samples producing large gradients. This method has been shown to reduce the accuracy disparity across groups on tabular datasets. It was also noted that computing different clipping bounds leaks additional information when compared to classical DP-SGD, and thus requires larger perturbations. This causes an additional trade-off, since, as shown in Section \ref{sec:issue_learning}, the noise magnitude is an important source of disparate impacts in DP-SGD. 
A different attempt uses early stopping \cite{zhang2021balancing}. The authors note that 
the number of training iterations is a crucial factor to balance utility, privacy, and fairness. Importantly, this method relies on the availability of a public validation set. 
Finally, motivated by the observation that the excessive risk across groups can be different during private training (see Section \ref{sec:issue_learning}), Tran et al.~\shortcite{tran2021differentially} suggest to add a targeted fairness constraint to the empirical risk minimizer. The constraint equalizes the difference among the groups' excessive risks. This simple solution was shown to reducing the excessive risk differences among groups while retaining high accuracy. 

\smallskip
The works above all protect each data sample in a dataset. The question regarding the necessity to protect exclusively the group attributes (e.g., gender or race) was first posed by Jagielski et al.~\shortcite{jagielski2019differentially}. Under this more permissive privacy setting, the authors propose two algorithms that balance privacy and equalized odds impacts. The first is a DP version of the post-processing method of Hardt et al.~\shortcite{hardt2016equality}, which {uses different decision thresholds for different groups to remove the disparate mistreatment.}. The second is a DP version of the method suggested by Agarwal et al.~\shortcite{agarwal2018reductions}, which augments the loss function with a penalizer that accounts for the reported fairness violations. 
While innovative, these algorithms require very large privacy budgets, which is partly due to the use of advanced composition to derive the final privacy loss. In a similar setting, Mozannar et al.~\shortcite{mozannar2020fair} introduced a simple yet effective solution: It first applies randomized response to protect the sensitive group labels. Then, it uses the penalizer model introduced in \cite{agarwal2018reductions} to obtain a good tradeoff between privacy and fairness. 
Lastly, Tran et al.~\shortcite{tran2020differentially} study a private extension of a Dual Lagrangian framework applied to learning tasks \cite{fioretto2020lagrangian}, in which the primal and dual steps are computed privately and privacy computation relies on the moment accountant \cite{AbadiCGMMT016}. While good privacy/fairness trade-offs are reported, this method comes at a steeper computational cost due to the  computations required by the private dual step. 

In the context of semi-supervised teacher ensemble models, Tran et al.~\shortcite{tran2021fairness} observes that the voting process of the teacher ensemble is subject to robustness issues especially in low voting confidence regimes (small perturbations may significantly affect the result of the voting result). To mitigate this issue, the authors explore the use of soft-labels in the voting ensemble. Teachers using soft labels report confidence score associated with each target label, rather than reporting solely the label with the largest confidence.
This additional information is carried at no additional privacy cost and it was shown helpful in achieving better privacy/fairness trade-offs. 

\smallskip
Finally, some recent solution tackling privacy and fairness has also arisen in the context of federated learning. 
Notably, Abay et al.~\shortcite{abay2020mitigating} 
proposed several pre-processing and in-processing bias mitigation solutions to improve fairness without affecting data privacy.
Finally, Padala et al.~\shortcite{padala2021federated} proposed a two-phase training step performed by each client. Clients first train a non-private model which maximizes accuracy while controlling the fairness violations. Then, they train a private model using DP-SGD to mimic the first, fair model. The updates obtained by this private model are thus broadcasted to an aggregator at each iteration.

\section{Challenges and Research Directions}
The current research at the intersection of differential privacy and fairness has shown promise in building solutions to realize more trustworthy systems. Furthermore, the analysis of the disparate impacts arising in several learning and decision tasks has paved the way to develop promising mitigating strategies. 
Despite these encouraging results, a number of challenges must be addressed to have a full understanding of the trade-offs between privacy, fairness, and accuracy.
{\bf (1)} The development of a unified theoretical framework to characterize and reason about fairness issues arising in general decision tasks is still missing. Of particular importance would be to capture the relation between the privacy loss values and the fairness violations resulting in both decision-making and learning settings. 
{\bf (2)} While the current focus in the analysis of fairness in private ML tasks has focused on data and algorithmic properties, it has also been observed that batch-size and learning rate may affect the Hessian spectrum of a network classifier \cite{yao2018hessian}. These observations may suggest that fairness in private ML tasks may be impacted by key hyper-parameters, including batch size, learning rates, and the depth of neural networks.
{\bf (3)} Another aspect that has been observed repeatedly when connecting privacy and fairness is their link with model robustness. While this observation arises both in decision and in learning tasks, an understanding of this link is currently missing.
{\bf (4)} A further important direction is the study of the disparate impacts that may arise in algorithms and generative models producing private synthetic datasets as well the development of mitigation measures. 
{\bf (5)} Finally, the development of software library to facilitate auditing fairness and bias issues in a private decision or learning task would be crucial to broaden the knowledge and adoption of these important issues.

Although the approaches surveyed are still in an early stage of their development, understanding the intricacies at the intersection of privacy, fairness, and accuracy will help shed light on the design of fairer ML systems and decision problems that use sensitive data. In turn, this will provide novel and unique perspectives for users and policymakers about the societal consequences of using differential privacy for critical processes, including predictions and decisions tasks.

\section*{Acknowledgement}
This research is partially supported by NSF grant 2133169, CUSE grant II-37-2021, a Google Scholar Research award and an Amazon Research Award. Its views and conclusions are those of the authors only.

\bibliographystyle{named}
\bibliography{lib.bib}

\end{document}